\documentclass[a4paper,twoside]{article}
\usepackage{geometry}
\newcommand{\smallref}[1]{{\fontsize{9}{11}\selectfont \ref{#1}}} 
\newcommand{\smalleqref}[1]{{\fontsize{9}{11}\selectfont \eqref{#1}}} 

\newcommand\blfootnote[1]{
    \begingroup
    \renewcommand\thefootnote{}\footnote{#1}
    \addtocounter{footnote}{-1}
    \endgroup
} 

\usepackage{epsfig}
\usepackage{subcaption}
\usepackage{calc}
\usepackage{amssymb}
\usepackage{amstext}
\usepackage{amsmath}
\usepackage{amsthm}
\usepackage{graphicx}
\usepackage{float}
\usepackage{pgfplots}
\usepackage{multicol}
\usepackage{pslatex}
\usepackage{apalike}
\usepackage{algorithm2e}
\usepackage[bottom]{footmisc}
\usepackage{SCITEPRESS}

\begin{document}

\setcounter{secnumdepth}{3} 

\renewcommand{\thesubsubsection}{\Alph{subsubsection}} 

\title{CuRLA: Curriculum Learning Based Deep Reinforcement Learning For Autonomous Driving}

\author{\authorname{Bhargava Uppuluri\sup{1,2}, Anjel Patel\sup{1,2}, Neil Mehta\sup{1,2}, Sridhar Kamath\sup{1,3} and Pratyush Chakraborty\sup{1,3}}
\affiliation{\sup{1}Birla Institute of Technology \& Science, Pilani, Hyderabad Campus, Jawahar Nagar, Kapra Mandal, Medchal District - 500078, Telangana, India}
\sup{2}\email{\{f20191279h, f20190126h, f20190177h\}@alumni.bits-pilani.ac.in, \sup{3}\{f20200976, pchakraborty\}@hyderabad.bits-pilani.ac.in}
}

\keywords{Computer Vision, Deep Reinforcement Learning, Variational Autoencoder, Proximal Policy Optimization, Curriculum Learning, Autonomous Driving.}

\abstract{In autonomous driving, traditional Computer Vision (CV) agents often struggle in unfamiliar situations due to biases in the training data. Deep Reinforcement Learning (DRL) agents address this by learning from experience and maximizing rewards, which helps them adapt to dynamic environments. However, ensuring their generalization remains challenging, especially with static training environments. Additionally, DRL models lack transparency, making it difficult to guarantee safety in all scenarios, particularly those not seen during training. To tackle these issues, we propose a method that combines DRL with Curriculum Learning for autonomous driving. Our approach uses a Proximal Policy Optimization (PPO) agent and a Variational Autoencoder (VAE) to learn safe driving in the CARLA simulator. The agent is trained using two-fold curriculum learning, progressively increasing environment difficulty and incorporating a collision penalty in the reward function to promote safety. This method improves the agent’s adaptability and reliability in complex environments, and understand the nuances of balancing multiple reward components from different feedback signals in a single scalar reward function.}

\onecolumn \maketitle \normalsize \vfill

\section{\uppercase{Introduction}}
\label{sec:introduction}
\blfootnote{P. Chakraborty and S. Kamath are with the Birla Institute of Technology \& Science (BITS) Pilani, Hyderabad Campus. B. Uppuluri, A. Patel, and N. Mehta were with BITS Pilani, Hyderabad Campus, and are now alumni.} The quest for safer, more efficient, and accessible transportation has made autonomous vehicles (AVs) a key technological innovation. Rule-based systems showed some promise~\cite{moravec1990sensor} but lacked the adaptability and robustness to tackle real-world scenarios. In the late 1900s, neural network-based supervised machine learning algorithms were trained on labeled datasets to predict steering angles, braking, and other control actions based on sensor data (\textit{e.g.}, ALVINN~\cite{NIPS1988_812b4ba2}). In the early 2000s SLAM techniques such as LiDAR and Radar improved understanding of vehicle position and surroundings, improving vehicle navigation accuracy~\cite{book1}, even in dynamic lighting and weather conditions -~\cite{article2}. The DARPA Grand Challenge in 2005 -~\cite{enwiki:1224665446} which saw Stanley emerge as the winner, marked significant milestones in autonomous driving. Stanley utilized machine learning techniques to navigate the unstructured environment, thus recognizing machine learning and artificial intelligence as essential components of autonomous driving technology.~\cite{Grigorescu_2019}

Eventually, with the advancement in camera technology, vision-based navigation became an area of research, with algorithms processing visual data to identify lanes, obstacles, and traffic signals. Although this approach worked very well in controlled environments, it faced challenges in dynamic scenarios and varying light conditions~\cite{dickmanns1987integrated}. Combining the advancements in Deep Learning and Computer Vision, CNNs were used to extract feature maps from visual data from the camera mounted on the vehicle. This led to breakthroughs in object detection, lane understanding, and road sign recognition (\textit{e.g.}, ImageNet)~\cite{article3}. Despite progress, fully automating decision-making and vehicle control in dynamic environments remains challenging, requiring significant manual feature engineering in ML and DL methods.

Here is where reinforcement learning comes into play, showing great potential for decision-making and controlling tasks. Combining DL with RL, where agents learn through trial and error in simulated environments, has significantly improved AV decision-making -~\cite{mnih2013playing}. Unlike rule-based systems, DRL agents can learn to navigate through diverse scenarios by trial and error behavior. This kind of learning also allows the agents to master intricate maneuvers and handle unexpected situations.~\cite{mnih2013playing} used DQN to showcase driving policy learning from raw pixel inputs in austere video game environments. However, a simple DQN ~\cite{mnih2013playing} will not work well in real-life applications, such as driving, as the action space is not discrete but continuous.

Deep Deterministic Policy Gradient (DDPG) is an on-policy actor-critic algorithm specifically designed to handle continuous action spaces by directly parameterizing the action-value function~\cite{lillicrap2019continuous}. The authors of~\cite{kendall2018learning} used this algorithm to drive a full-sized autonomous vehicle. The system was first trained in simulation before being introduced in real-time using onboard computers. It used an actor-critic model to output the steering angle and speed control with the help of an input image. They suggested having a less sparse reward function and using Variational Auto Encoders (VAEs)~\cite{kingma2022autoencoding} for better state representation. DDPG also requires many interactions with the environment, making the training slow, especially in a high-dimensional and complex environment. DDPG training can also be unstable, especially in sparse reward or non-stationary environments. This instability can manifest as oscillations or divergence during training ~\cite{mnih2016asynchronous}. Proximal Policy Algorithm (PPO)~\cite{schulman2017proximal} was developed to address these issues. 

In our work, we use PPO~\cite{schulman2017proximal} and Curriculum Learning~\cite{curriculumlearning} for the self-driving task. To obtain a better representation of the state, we have used Variational Auto Encoders (VAE)~\cite{kingma2022autoencoding} to encode the current scene from CARLA~\cite{dosovitskiy2017carla}, the urban driving simulator (Town 7). Our paper builds upon the foundational work about accelerated training of DRL-based autonomous driving agents presented in~\cite{Vergara2019}. Salient features of our work:
\begin{itemize}
    \item Introduction of Curriculum learning in the training process allows the agent to learn the easier tasks like moving forward initially and as difficulty increases to make it learn more difficult tasks, like maneuvering in traffic or avoiding high-speed collisions.
    \item We have introduced a refined reward function that gives a higher reward to the agent to travel at higher speeds. This is important to increase the average speed, reducing travel time.
    \item Unlike our base paper, our reward function takes into account the collision penalty as well as other rewards like angle, centering, and speed reward. This is crucial to make the reward function less sparse and aid a smoother driving experience.
    \item The combination of the curriculum learning approach, involving increasing traffic density and augmenting the reward function, as well as the modified reward function, helps in the faster training process and better average speed of the agent. 
\end{itemize}
We name this method CuRLA - Curriculum Learning Based Reinforcement Learning for Autonomous Driving, as curriculum learning is integral to the features in our work. These features and the improvements they bring about will be discussed in the paper in further detail.

\section{\uppercase{Preliminaries}}
In this section, we provide the foundational tools, concepts, and definitions necessary to understand the subsequent content of this paper. We begin by introducing the environment used for training, followed by a brief introduction to Policy Gradient RL algorithms. Next, we discuss the concept of curriculum learning, and finally, we detail the encoder used in our approach.

\subsection{CARLA Driving Simulator}
The rise of autonomous driving systems in recent years owes much to the emergence of sophisticated simulation environments. CARLA~\cite{dosovitskiy2017carla} is a critical resource for researchers and developers in autonomous driving, providing an open-source, high-fidelity simulator. Its capabilities include realistic vehicle dynamics, sensor emulation (such as LiDAR, radar, and cameras), and dynamic weather and lighting conditions. Moreover, CARLA's scalability enables us to simulate large-scale scenarios involving multiple vehicles and pedestrians interacting in complex urban environments. Its standardized metrics and scenarios facilitate fair comparisons between different self-driving approaches in the field. We particularly chose the CARLA simulator as it also provides a collision intensity whenever a vehicle collides with another object in the environment, and we use this in the reward function design.

\subsection{Policy Gradient Methods}
Policy gradient methods~\cite{NIPS1999_464d828b} are pivotal in reinforcement learning for continuous action spaces, directly parameterizing policies, enabling the learning of complex behaviors. Unlike value-based approaches that estimate state/action values, these methods learn a probabilistic mapping from states to actions, enhancing adaptability in stochastic environments. These methods optimize policy parameters "\(\theta\)" in continuous action spaces through on-policy gradient ascent on the performance objective \(J(\pi_\theta)\).

Trust Region Policy Optimization (TRPO)~\cite{schulman2017trust} is a type of policy gradient method that stabilizes policy updates by imposing a KL divergence constraint~\cite{KLD}, preventing large updates. However, TRPO's complex implementation and incompatibility with models sharing parameters or containing noise are drawbacks.

Proximal Policy Optimization (PPO) ~\cite{schulman2017proximal} is an on-policy algorithm suited for complex environments with continuous action and state spaces. It builds upon TRPO~\cite{schulman2017trust} by using a clipped objective function for gradient descent, simplifying implementation with first-order optimization while maintaining data efficiency and reliable performance. In later sections, we will see how this has been implemented in our work.

\subsection{Curriculum Learning}
Curriculum Learning~\cite{curriculumlearning} is a strategy aimed at enhancing the efficiency of an agent's learning process by optimizing the sequence in which it gains experience. By strategically organizing the learning trajectory, either performance or training speed on a predefined set of ultimate tasks can be improved. By quickly acquiring knowledge in simpler tasks, the agent can leverage this understanding to reduce the need for extensive exploration to tackle more complex tasks.~\cite{narvekar2020curriculum}

\subsection{Variational Autoencoder}
Autoencoders~\cite{Rumelhart1986LearningIR} are essentially generative neural networks that comprise an encoder followed by a decoder, whose objective is to transform input to output with the least possible distortions~\cite{Baldi2011AutoencodersUL}. 

VAEs~\cite{kingma2022autoencoding} excel in reinforcement learning by producing varied and structured latent representations, enhancing exploration strategies, and adapting to novel states. Moreover, their probabilistic framework enhances resilience to uncertainty, which is crucial for adept decision-making in dynamic settings.

\begin{figure}[H]
\centering
        \includegraphics[width=\linewidth, height=6cm]{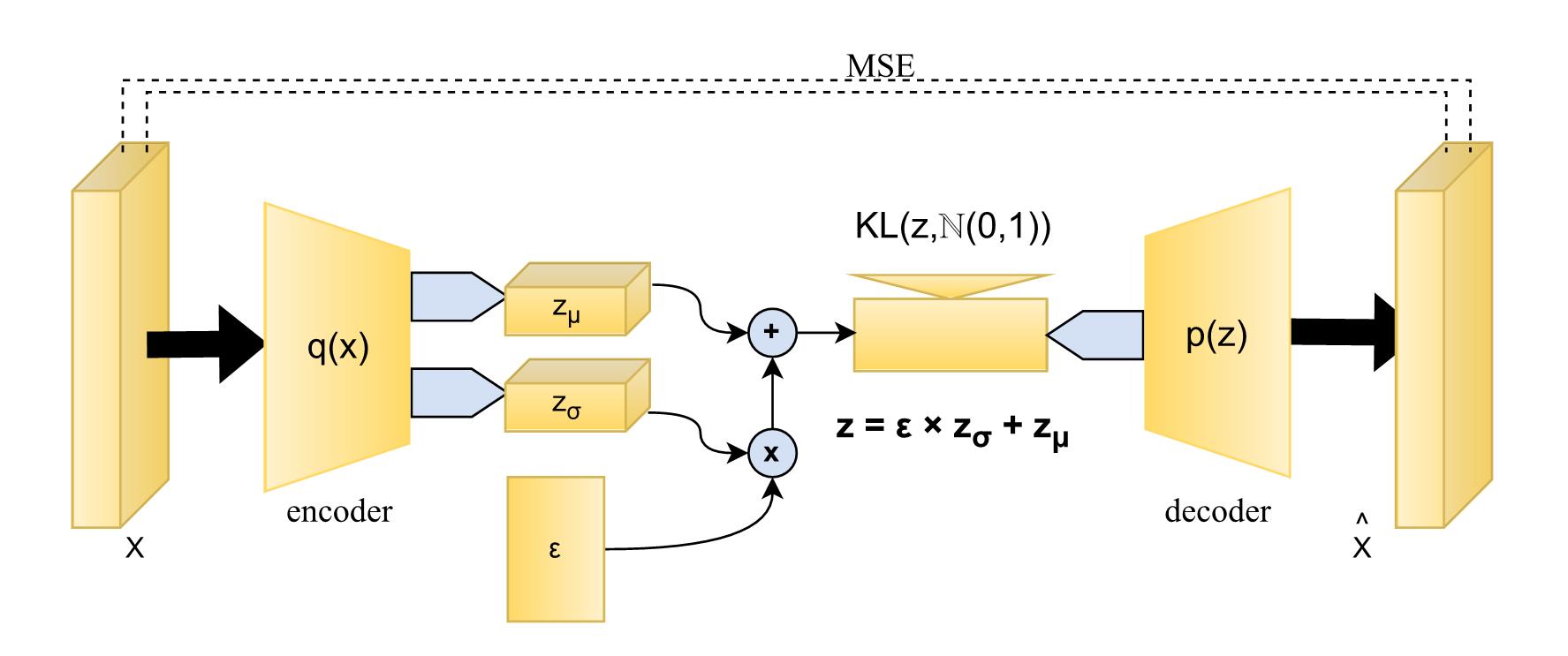}
        \caption{Variational autoencoder architecture.}
        \label{fig:VAE}
\end{figure}

Figure (\smallref{fig:VAE}) shows a variational autoencoder architecture. The encoder operates on the input vector, yielding two vectors, \(z_\mu\) and \(z_\sigma\). Then, a sample \(z\) is drawn from the distribution \(N(z_\mu; z_\sigma)\), which is fed into the decoder \(p\), producing a reconstructed signal.

\begin{figure}[H]
\centering
        \includegraphics[width=\linewidth, height=4.0cm]{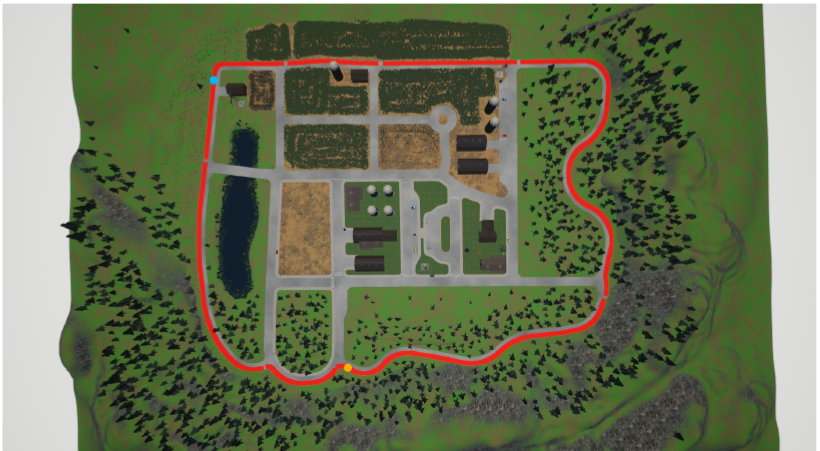}
        \caption{Top down view of the lap in Town 7.}
        \label{fig:CT7tv}
\end{figure}

\section{\uppercase{Experimental Study and Result Analysis}}

In this section, we present the experimental setup, methodology, and results of our study. We start by describing the experimental environment, followed by a detailed explanation of the evaluation metrics and the baseline methods for comparison. We then present the results of our experiments and finally discuss the implications of our findings and compare our results with existing work in the field.

\subsection{Proposed Model and Experimental Study}
Our model is named as CuRLA and the model used in the base paper is named as Self-Centering Agent (SCA) for further reference. We also perform experiments using only curriculum learning for the optimized reward function to compare it's performance to the two-fold curriculum learning method that is implemented in CuRLA. This model is named One-Fold CL. We use Town 7 from the list of Towns in CARLA Lap Environment~\cite{dosovitskiy2017carla} as shown in figure (\smallref{fig:CT7tv}). The reason for using this specific town is that it provides a highway environment without any exits, thus making our job of end-to-end driving easier. In our experiments, curriculum learning is employed by gradually increasing traffic volume and introducing functionalities to the agent in it's reward function in stages, allowing it to quickly grasp the basics of the environment and efficiently tackle more complex tasks. 

\begin{table}[H]
    \centering
    \caption{VAE Parameters.}
    \begin{tabular}{|c|c|}
    \hline
        \textbf{Hyperparameter}&{\textbf{Value}} \\
        \hline
        $z_{dim}$ & 64 \\
        Architecture & CNN \\
        Learning rate $\alpha$ & $1e-4$  \\    
        $\beta$ & 1 \\
        Batch size $N$ & 100\\
        Loss & BCE \\
        \hline
    \end{tabular}
    \label{tab:vaehyp}
\end{table}

The VAE~\cite{kingma2022autoencoding} we have used serves as feature extractors by compressing high-dimensional observations into a lower-dimensional latent space. This aids the learning process by providing a more manageable representation for the agent.  Variational Autoencoder (VAE) is chosen as it is used for learning probabilistic representations of the input in the latent space, unlike regular autoencoders~\cite{Rumelhart1986LearningIR}, which learn deterministic representations. We use the VAE architecture from~\cite{Vergara2019} to encode the state of the environment, utilizing the pre-trained VAE from the same study to replicate the encodings. The dataset used to train the VAE contains 9000 training and 1000 validation examples (total of \(10,000\) \(160\times80\) RGB images), all collected manually from the driving simulator environment. 

We have made use of the state-of-the-art PPO-Clip~\cite{schulman2017proximal} algorithm to optimize our agent’s policy and to arrive at an optimal policy that maximizes the return. PPO-Clip clips the policy to ensure that the new policy does not diverge far away from the old policy. The policy update equation of PPO-Clip is given as:

\begin{equation}
\theta_{k+1}=\arg\max_{\theta}E_{s,a\sim\theta_k}\left[L(s,a,\theta_k,\theta)\right]
\end{equation}
Here, L is given as:
\small
\begin{multline}
    L(s,a,\theta_k,\theta)=\\\min\left(\frac{\pi_\theta(a|s)}{\pi_{\theta_k}(a|s)} A^{\pi_{\theta_k}}
    ,clip\left(\frac{\pi_\theta(a|s)}{\pi_{\theta_k}(a|s)},1+\epsilon,1-\epsilon \right)A^{\pi_{\theta_k}} \right)
\end{multline}
\normalsize

Here, \(\theta\) refers to the policy parameters being updated, and \(\theta_k\) is the policy parameters currently being used in the \(k^{th}\) iteration to get the next iterations parameters, \(\theta_{k+1}\). Also, \(\epsilon\) is the clipping hyperparameter defining how much the new policy can diverge from the old one. The \(\min\) in the objective computes the minimum of the un-clipped and clipped objectives.

\begin{figure}[H]
\centering
        \fbox{\includegraphics[width=0.98\linewidth, height=5cm]{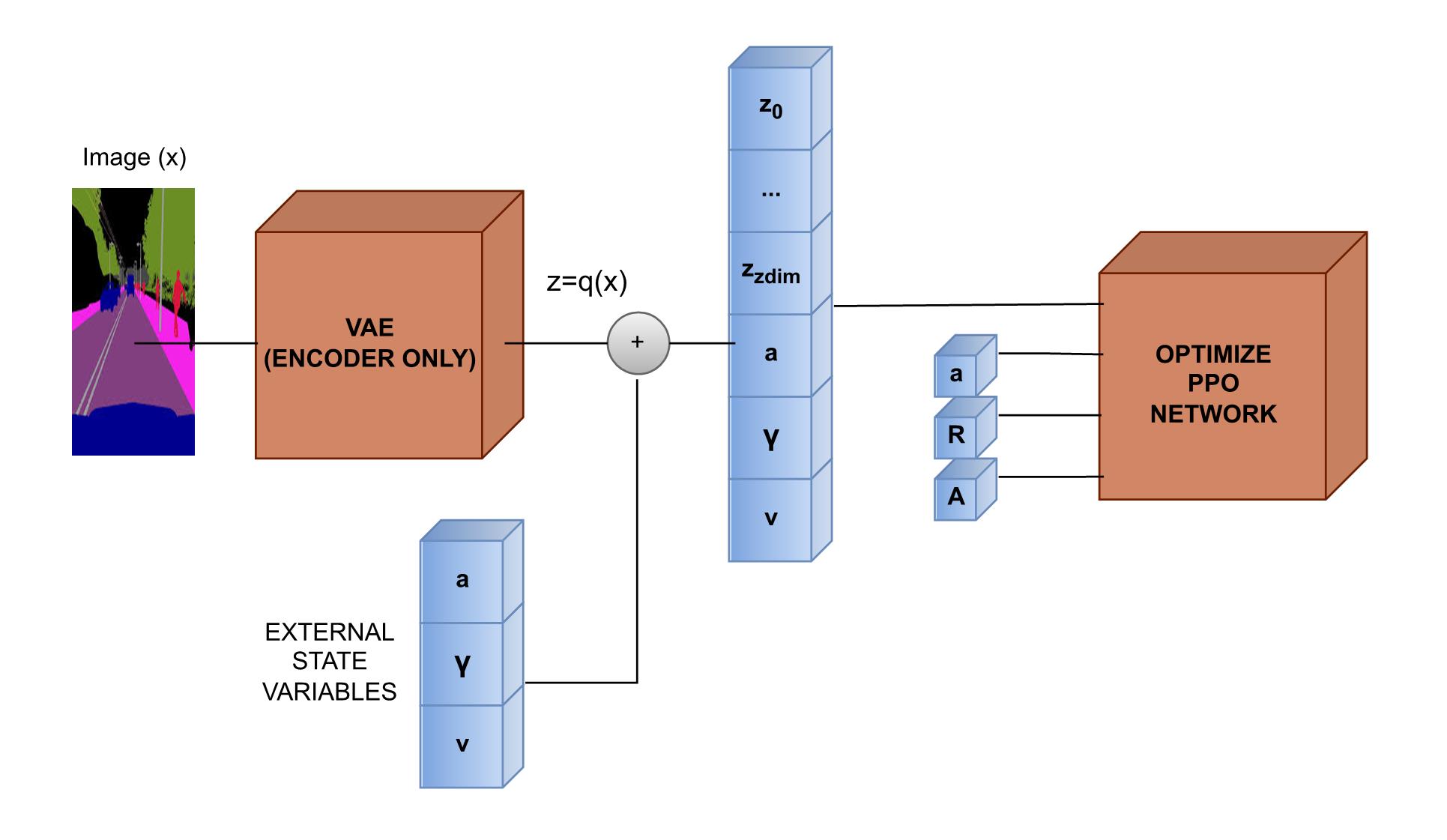}}
        \caption{PPO+VAE training architecture.}
        \label{fig:trainarch}
\end{figure}

The overall graphical representation of the PPO+VAE training process we have used in our paper is shown in the figure (\smallref{fig:trainarch}).In the figure (\smallref{fig:trainarch}), the external variables are acceleration, steering angle, and speed from top to bottom.

The models are trained with the same parameters mentioned in ~\cite{Vergara2019}. Table \smallref{tab:modhyp} shows the complete list of hyperparameters used in the experiments. All models are trained for 3500 episodes, and the evaluation is done once every 10 episodes. Curriculum learning is implemented in both CuRLA and One-Fold CL. After 1500 episodes, traffic and a collision penalty are introduced in CuRLA. In One-Fold CL, only the collision penalty is introduced after 1500 episodes, and traffic is present from the first episode itself. An episode ends when the agent finishes three laps, drifts off the center of the lane by more than 3 metres, or has a speed of less than 1 km/hr for 5 seconds or more. A collision will not lead to the end of an episode (unless the agent collides and maintains a low speed under 1 km/hr for 5 seconds and more due to the collision). The results we obtain are shown in Figs [\smallref{fig:train_distance_traveled},\smallref{fig:train_average_speed},\smallref{fig:eval_distance_traveled},\smallref{fig:eval_average_speed}].

\begin{table}[H]
    \centering
    \caption{Model Parameters.}
    \begin{tabular}{|c|c|}
    \hline
        \textbf{Hyperparameters}&{\textbf{Value}} \\
        \hline
        Batch size $M$ & 32 \\
        Entropy loss scale $\beta$ & 0.01 \\
        GAE parameter $\lambda$ & 0.95 \\
        Initial noise $\sigma_{init}$ & 0.4 \\
        Epochs $K$ & 3 \\
        Discount $\gamma$ & 0.99 \\
        Value loss scale $\alpha$ & 1.0\\
        Learning rate & $1e-4$  \\  
        PPO-Clip parameter $\epsilon$ & 0.2 \\
        Horizon $T$ & 128\\
        \hline
    \end{tabular}
    \label{tab:modhyp}
\end{table}

\subsection{Improvements Made}
\subsubsection{Reward Function Optimization}
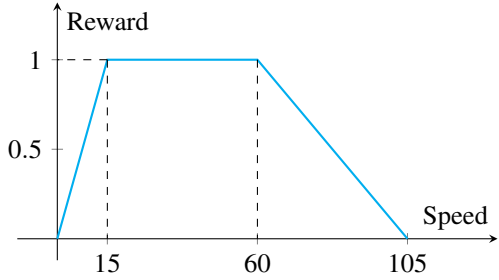
\begin{figure}[t]
    \centering
        \begin{tikzpicture}
            \begin{axis}[
                xlabel={Speed},
                ylabel={Reward},
                xmin=0, xmax=120,
                ymin=0, ymax=1.2,
                axis lines=middle,
                enlargelimits,
                xtick={0, 15, 60, 105},
                ytick={0, 0.5, 1},
                height=5cm,
                width=0.5\textwidth
            ]
            \addplot[
                color=cyan,
                thick,
                mark=none
            ]
            coordinates {
                (0,0) (15,1) (60,1) (105,0) 
            };  
            \draw[dashed] (axis cs:15,1) -- (axis cs:15,0);
            \draw[dashed] (axis cs:60,1) -- (axis cs:60,0);
            \draw[dashed] (axis cs:15,1) -- (axis cs:0,1);
            \end{axis}
        \end{tikzpicture}
        \caption{
            SCA Reward Function.
            }
        \label{fig: scarf}
\end{figure}

\begin{figure}[t]
    \centering
        \begin{tikzpicture}
            \begin{axis}[
                xlabel={Speed},
                ylabel={Reward},
                xmin=0, xmax=120,
                ymin=0, ymax=1.2,
                axis lines=middle,
                enlargelimits,
                xtick={0, 15, 60, 105},
                ytick={0, 0.5, 1},
                height=5cm,
                width=0.5\textwidth
            ]
            \addplot[
                color=cyan,
                thick,
                mark=none
            ]
            coordinates {
                (0,0) (15,0.5) (60,1) (105,0) 
            };
            \draw[dashed] (axis cs:15,0.5) -- (axis cs:15,0);
            \draw[dashed] (axis cs:60,1) -- (axis cs:60,0);
            \draw[dashed] (axis cs:15,0.5) -- (axis cs:0,0.5);
            \draw[dashed] (axis cs:60,1) -- (axis cs:0,1);
            \end{axis}
        \end{tikzpicture}
        \caption{CuRLA \& One-Fold CL Reward Function.}
        \label{fig: hcv6rf}
\end{figure}
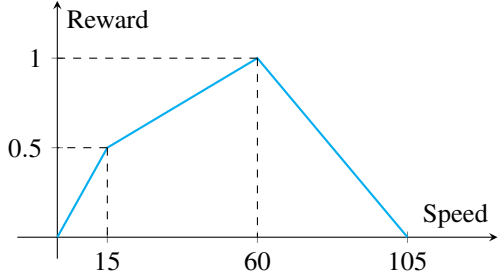

Keeping the architecture of the VAE and the RL algorithm the same as the original paper, we chose to change the original reward function. The original reward function \smalleqref{eq:sca} has 3 components: the angle reward, the centering reward, and the speed reward. Our revised reward function \smalleqref{eq:hcv6} has 4 components, the three original components, and we have also added a collision penalty, to encourage the agent against unsafe driving behaviour.
\begin{equation}
    \label{eq:sca}
    r = r_{\alpha} \cdot r_{d} \cdot r_{v}
\end{equation}
\begin{equation}
    \label{eq:hcv6}
    r^{\prime} = r_{\alpha} \cdot r_{d} \cdot r_{v^{\prime}} + r_{c}
\end{equation}
\begin{enumerate}
    \item {\textbf{Angle Reward:} The angle reward component ensures that the agent is aligned with the road. Angle $\alpha$, determines the angle difference between the forward vector of the vehicle (the direction the vehicle is heading in) and the forward vector of the vehicle waypoint (the direction the waypoint is aligned with). Using $\alpha$, the angle reward $r_{\alpha}$ is defined by \begin{equation}
    r_{\alpha} =  \max\left(1 - \left| \frac{\alpha}{\alpha_{max}}\right|, 0\right)
    \label{eq:angle_reward}
    \end{equation}
    \vspace{2pt}
    where $\alpha_{max} = 20^{\circ}$ ($\pi/9$ radians).
    
    This ensures that the angle reward  $r_{\alpha} \in [0,1]$, and that the angle reward linearly decreases from 1 (perfect alignment) to 0 (when the deviation is 20$^\circ$ or more).}
    \item {\textbf{Centering Reward:} The centering reward factor ensures that the driving agent stays to the center of the lane. The usage of a high-fidelity simulator like CARLA enables us to have a precise measurement of distance between objects in the environment. To reward the agent for staying within the center of the lane, we use the distance $d$ between the center of the car and the center of the lane to define the centering reward by
    \begin{equation}
        r_{d} = \left(1 - \frac{d}{d_{max}} \right)
    \end{equation}
    where $d_{max} = 3$.

    This ensures that the centering reward component $r_{d} \in [0,1]$ (as the episode terminates when the agent goes off center by more than 3 metres) and that $r_{d}$ is inversely proportional to $d$, encouraging more centering for the agent while driving.}
    \item {\textbf{Speed Reward:} While keeping the angle and centering reward the same for all three agents, we change the speed reward component. The minimum speed $v_{min}$, maximum speed $v_{max}$, and target speed $v_{target}$ are taken as 15 Km/hr, 105 Km/hr and 60 Km/hr respectively, and are kept as the same for both the original and updated speed reward components. The original speed reward component $r_{v}$ and the new speed reward component $r_{v^{\prime}}$ are then defined by
    \begin{equation}
        \label{eq:sca_speed}
        r_{v} =
        \begin{cases}
            \frac{v}{v_{min}}, & v < v_{min}, \\
            1, & v_{min} \leq v \leq v_{target}, \\
            1 - \frac{v - v_{target}}{v_{max} - v_{target}}, & v_{target} < v \leq v_{max}
        \end{cases}
    \end{equation}
    \vspace{2pt}
    \begin{equation}
        \label{eq:hcv6_speed}
        r_{v^{\prime}} =
        \begin{cases}
            0.5 \cdot \frac{v}{v_{min}}, & v < v_{min}, \\
            1 - \frac{v_{target} - v}{v_{target} - v_{min}}, & v_{min} \leq v \leq v_{target}, \\
            \frac{v_{max} - v}{v_{max} - v_{target}}, & v_{target} < v \leq v_{max}
        \end{cases}
    \end{equation}

    where $v$ is the current speed of the agent.
    
    As seen in the graphs (Fig.\smallref{fig: scarf} \& Fig.\smallref{fig: hcv6rf}), in the original speed reward function (Fig.\smallref{fig: scarf}), the graph is constant in the range [$v_{min}$, $v_{target}$] and receives the highest reward of 1. This is misleading as the agent can get confused, deciphering the minimum speed as the target speed itself, as it is getting a constant reward of 1 at either speed. This ensures that the agent will drive slower, as ensuring a higher centering and angle reward at a lower speed is much easier than at a higher speed. To rectify this, we have replaced the constant graph in [$v_{min}$, $v_{target}$] with an increasing function (see Fig.\smallref{fig: hcv6rf}) to prioritize getting as close to the target speed as possible without losing too much performance on the angle and centering reward components. Both CuRLA and One-Fold CL use this revised reward function, whereas SCA uses the original reward function.}

    \item {\textbf{Collision Penalty:} A collision penalty factor was introduced for both One-Fold CL and CuRLA to ensure the agent explicitly learns the behaviour of safe driving, avoiding collisions with other objects and vehicles in the environment. The advantage of using a simulator like CARLA also enables us to get a collision intensity ($I_{c}$) value between the agent and other objects in the environment, which we use to devise the collision penalty. The collision penalty $r_{c}$ is defined by
    \begin{equation}
        \label{eq:collisionpenalty}
        r_{c} = \max(-1, -\log_{10}(\max(1, I_{c})))
    \end{equation}

    This penalty ensures $r_{c} \in [-1, 0]$, thus ensuring $r^{\prime} \in [-1, 1]$.
    }
\end{enumerate}


\subsubsection{Curriculum Learning}
In CuRLA, curriculum learning is implemented in a two-fold manner. Firstly, we gradually increase the traffic volume of our simulation environment as the number of episodes increases. The agent gets to focus on learning to drive in a traffic-free environment initially and then slowly navigate through traffic in the later epochs. Secondly, the functionalities of the agent are gradually added. As the agent learns to drive in a traffic-free environment, we introduce a collision penalty while increasing the volume of traffic to teach it to avoid collisions with other vehicles. The reasoning behind adding this collision penalty is that while the current reward function accounts for smooth driving, it does not punish the agent enough for colliding with the other vehicles. All agents have been trained with traffic included in the roads, making a collision penalty extremely important. This method of training helped the agent to learn the basics of the environment quickly and enabled it to learn harder tasks efficiently while also keeping safety as a factor. 

\subsection{Metrics and Result Analysis}
The models are compared on two metrics - distance travelled and average speed. These metrics are chosen as they capture an accurate representation of the trade-off between efficiency and performance in autonomous driving scenarios. The distance traveled metric emphasizes the model's capability to maximize the length of the path it traverses, reflecting its ability to sustain long journeys without interruption. Conversely, the average speed metric accounts for both the distance covered and the time taken to complete the journey, offering a thorough assessment of the model's performance in terms of both speed and efficiency. Evaluating these metrics provides a detailed understanding of how each model balances speed and distance, which are essential factors in assessing the overall effectiveness of autonomous driving systems. The evaluation of our models against the base model provided insightful findings. (All graphs have been considered with a smoothing factor of 0.999)

\begin{figure*}[t]
    \begin{minipage}[t]{0.48\textwidth}
        \centering
        \includegraphics[width=\textwidth, height=5cm]{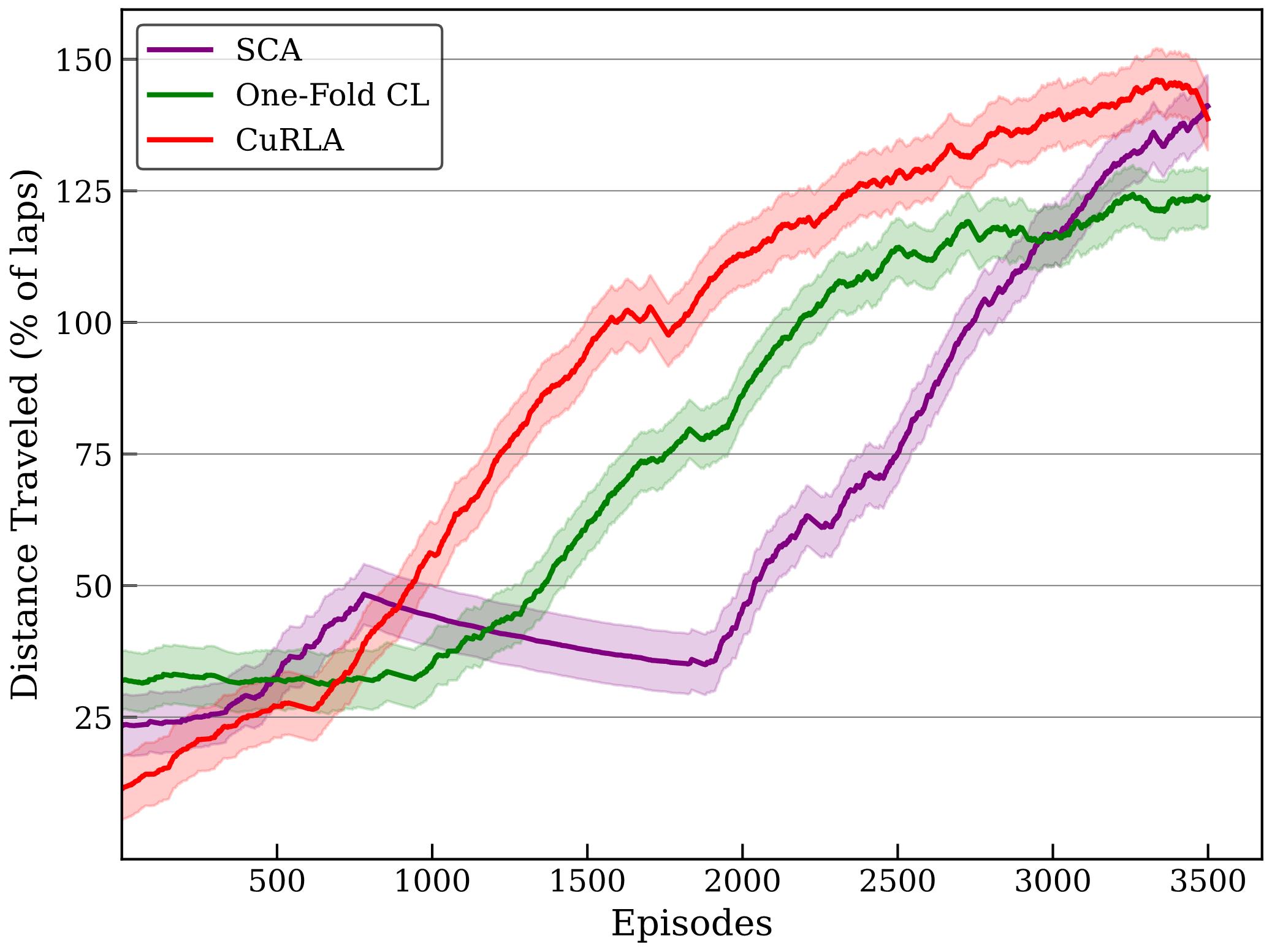}
        \caption{Training Metric: Distance Traveled.}
        \label{fig:train_distance_traveled}
    \end{minipage}
    \hfill
    \begin{minipage}[t]{0.48\textwidth}
        \centering
        \includegraphics[width=\textwidth, height=5cm]{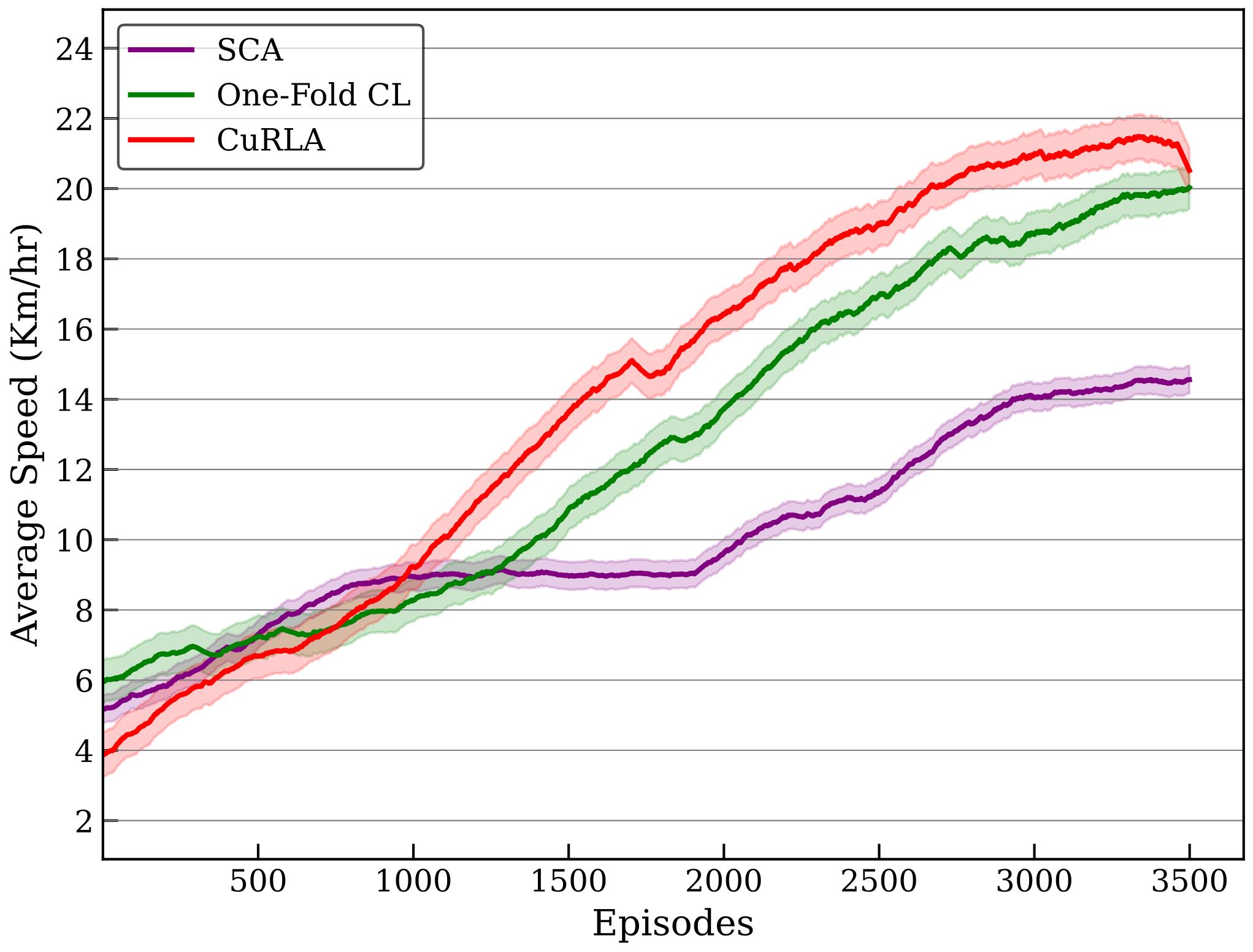}
        \caption{Training Metric: Average Speed.}
        \label{fig:train_average_speed}
    \end{minipage}
    
    \vspace{1em}
    
    \begin{minipage}[t]{0.48\textwidth}
        \centering
        \includegraphics[width=\textwidth, height=5cm]{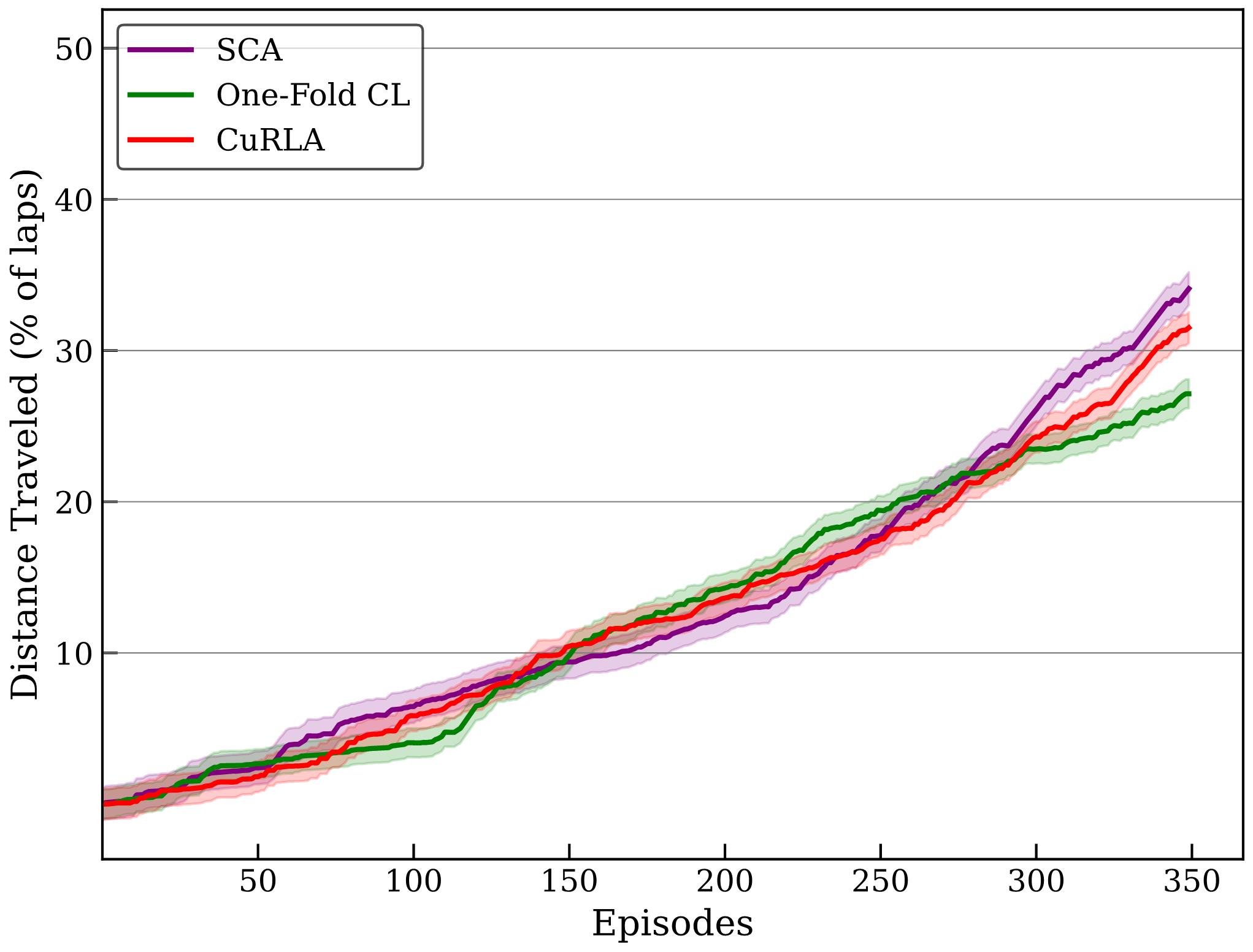}
        \caption{Evaluation Metric: Distance Traveled.}
        \label{fig:eval_distance_traveled}
    \end{minipage}
    \hfill
    \begin{minipage}[t]{0.48\textwidth}
        \centering
        \includegraphics[width=\textwidth, height=5cm]{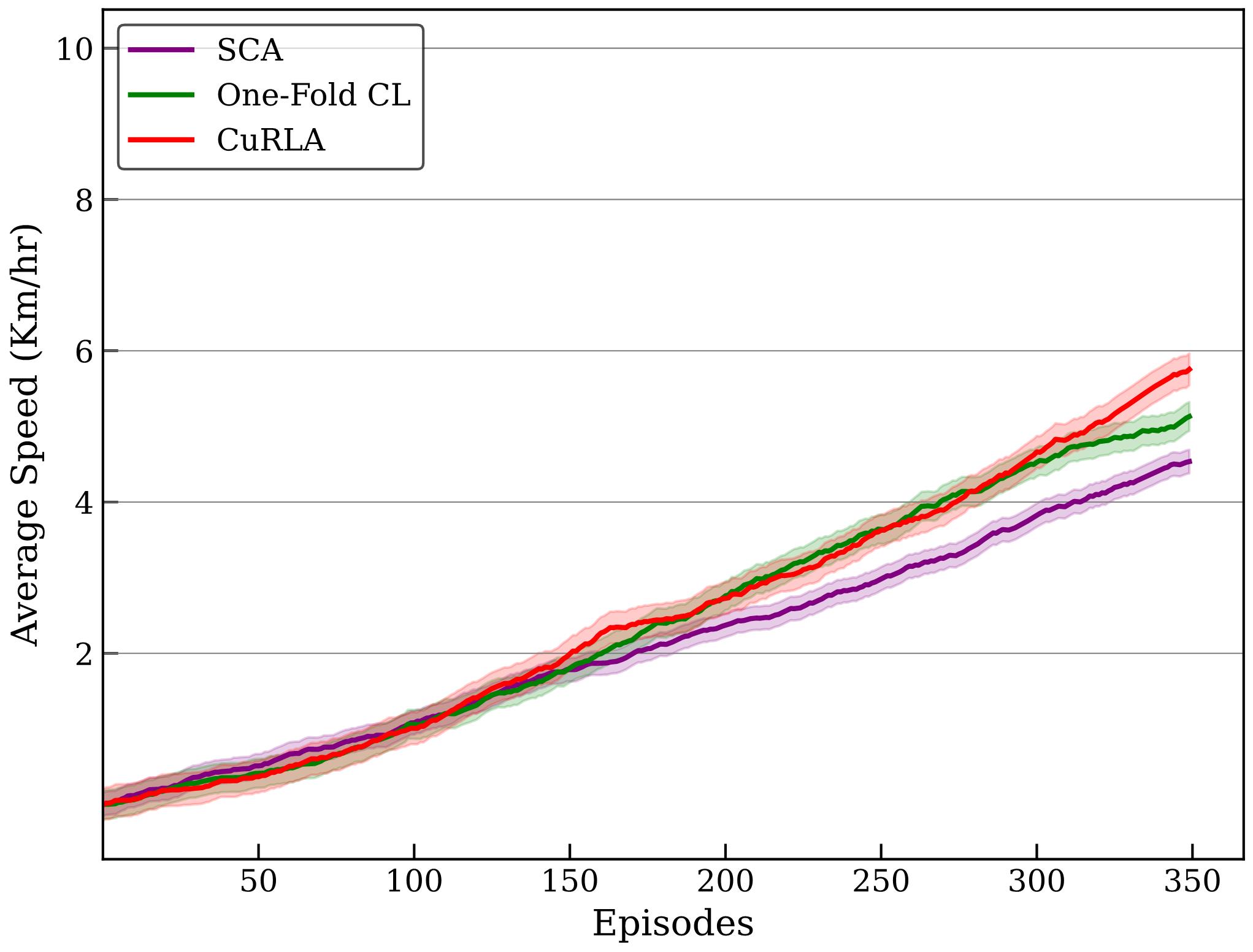}
        \caption{Evaluation Metric: Average Speed.}
        \label{fig:eval_average_speed}
    \end{minipage}
\end{figure*}

In our assessment, for the metric of Distance Traveled, we measure it by the number of laps traveled by the vehicle, where 100\% is considered as one lap finished. As it can be seen in Fig.\smallref{fig:train_distance_traveled} \& Fig.\smallref{fig:eval_distance_traveled}, CuRLA and SCA have similar performance, completing approximately 1.5 laps on average in training and 0.3 laps in evaluation. Meanwhile, One-Fold CL did worse than both CuRLA and SCA, completing approximately 1.25 laps in training and 0.25 laps in evaluation. CuRLA does perform slightly better in training compared to SCA, which performs slightly better in evaluation. However, from the graph it can be seen that CuRLA had a consistent learning curve while training compared to SCA, which dropped in performance and then picked back up. One-Fold CL also performed comparably to both models, albeit slightly worse. This attests to the improvement in performance that two-fold curriculum learning shows over simple curriculum learning during training and training without curriculum learning.

The assessment of the metric of Average Speed highlighted the strengths of our assumption in changing the underlying speed reward component. During training, as can be seen in Fig.\smallref{fig:train_average_speed}, both CuRLA and One-Fold CL significantly outperform SCA reaching average speeds of 22 Km/hr and 20 Km/hr in training, compared to SCA's 14 Km/hr. The difference is not as much during evaluation (Fig.\smallref{fig:eval_average_speed}), but CurLA and One-Fold CL still outperform SCA here as well, with CuRLA reaching an average speed of 6 Km/hr and One-Fold CL reaching an average speed of 5 Km/hr, compared to SCA's 4 Km/hr. This superior performance of CuRLA and One-Fold CL (on the revised reward function) compared to the SCA agent (using the original reward function) underscores our reward function's efficiency in optimizing speed-related aspects.  

\section{\uppercase{Conclusions}}
\label{sec:conclusion}
In this paper, we presented a model (CuRLA) that used a PPO+VAE architecture and two-fold curriculum learning along with a reward function tuned to accelerate the training process and achieve higher average speeds in autonomous driving scenarios. We show the performance of two-fold curriculum learning against simple curriculum learning (One-Fold CL agent), as well as the performance of agents on the revised reward function compared to the base reward function. While CuRLA and One-Fold CL perform comparably to the base agent (SCA) in the distance traveled metric (with CuRLA performing slightly better and One-Fold CL being slightly worse), a significant improvement in average speed is observed. This prioritization of speed was a deliberate design choice. The distance traveled metric solely optimizes for maximizing the traversed path length. Conversely, the average speed metric inherently optimizes for both distance and the time taken to complete the journey, effectively accounting for two performance factors within a single measure. The performance of CuRLA and One-Fold CL agents, when compared to SCA, also attest to the benefits of using curriculum learning while training, and how decomposing the tasks in the autonomous driving problem helps agents to learn better and faster. Integrating multiple objectives into a single scalar reward function often leads to suboptimal agent performance. However, by employing curriculum learning during training, we can enable agents to master the nuances of each reward component more effectively. This approach facilitates better understanding of the environment and objectives, and ultimately enhances overall performance. 

Future research will focus on enhancing performance by updating the architecture and algorithms. One area of interest is investigating vision-based transformers and advanced transformer-based reinforcement learning methods for autonomous driving control. This entails replacing the current Variational Autoencoder with architectures like Vision Transformers (ViT, Swin Transformer, ConvNeXT) tailored for raw visual data. Furthermore, newer techniques such as Decision Transformers or Trajectory Transformers could replace the Proximal Policy Optimization (PPO) algorithm to potentially enhance decision-making capabilities. Another promising area for future research is Multi-Objective Reinforcement Learning (MORL)~\cite{van2014multi,DBLP:journals/corr/abs-2103-09568,6918520}, where an agent optimizes multiple reward functions, each representing different objectives. Evaluating these advancements through simulated testing may lead to substantial performance improvements.

\bibliographystyle{apalike}
{\small
\bibliography{CuRLA}}

\end{document}